\documentclass[twocolumn]{article}
\usepackage{graphicx}
\usepackage{caption}
\usepackage{booktabs}
\usepackage{multirow}
\usepackage{longtable}
\usepackage{geometry}
\usepackage{lscape}
\usepackage{hyperref}
\usepackage{fancyref}
\usepackage{amsmath, amssymb, amsthm}

\theoremstyle{plain}
\newtheorem{theorem}{Theorem}[section]
\newtheorem{proposition}[theorem]{Proposition}

\newtheorem{conjecture}[theorem]{Conjecture}

\theoremstyle{definition}
\newtheorem{definition}[theorem]{Definition}
\newtheorem{example}{Example}

\theoremstyle{remark}

\newcommand{\RR}{\mathbb{R}}

\usepackage[natbib=true,style=numeric-comp,backend=bibtex,useprefix=true]{biblatex}
\title{Cardinality augmented loss functions}
\author{Miguel O'Malley\thanks{Max Planck Institute for Mathematics in the Sciences, ScaDS.AI Institute of Universitat Leipzig, \texttt{miguel.omalley@mis.mpg.de}}}

\begin{document}

\maketitle

\begin{abstract}
    Class imbalance is a common and pernicious issue for the training of neural networks. Often, an imbalanced majority class can dominate training to skew classifier performance towards the majority outcome. To address this problem we introduce cardinality augmented loss functions, derived from cardinality-like invariants in modern mathematics literature such as magnitude and the spread. These invariants enrich the concept of cardinality by evaluating the `effective diversity' of a metric space, and as such represent a natural solution to overly homogeneous training data. In this work, we establish a methodology for applying cardinality augmented loss functions in the training of neural networks and report results on both artificially imbalanced datasets as well as a real-world imbalanced material science dataset. We observe significant performance improvement among minority classes, as well as improvement in overall performance metrics.
\end{abstract}

\section{Introduction}

In training procedures for neural networks, often problems arise where one class is overrepresented against other classes in the training dataset, leading to bias in optimization. For example, a classifier with a 90\% majority class can naively achieve 90\% accuracy by always guessing the majority class. While resampling and reweighting schemes through scoring regimes are possible responses to class imbalance, these solutions often leave much to be desired through arbitrarily chosen parameters and reliance on the training set's balance.

We introduce the usage of cardinality-like invariants for the purpose of computing loss in neural networks. Cardinality-like invariants augment traditional cardinality, in that they evaluate the `effective cardinality' of a metric space. For instance, magnitude, one such invariant we evaluate and utilize in this work, is often described as the `effective diversity' of a metric space. The original motivation for the study of magnitude originates in the work of Solow and Polasky~\cite{solowandpolasky}, wherein the authors proposed the invariant as a method of evaluating the diversity of natural ecosystems. Magnitude's formal establishment and much of the theoretical foundation for its study is attributable to Leinster's work~\cite{magnitude}.

We apply cardinality-like invariants as loss functions in a manner similar to an averaging method for loss over a batch. Instead of taking the average, sum, or maximum over loss observations, we apply a cardinality-like invariant to determine the unique observed loss. The idea of this application is to balance loss towards unique mistakes instead of overcorrecting towards batches where the model makes identical errors. If, for example, a classification has an issue where dogs are often misclassified as foxes and happens to receive a batch of mostly dogs, the repeated fox errors will cause other error evaluations to spike but will appear as effectively one error for a cardinality-like invariant. 

\section{Related work}

Andreeva et. al. demonstrate that magnitude is directly correlated to test accuracy for neural networks~\cite{magnitudenn}, when evaluated on the space of models for a given model space. While this correlation is well substantiated, it is difficult to implement given the immense size of modern parameter spaces. Our methodology applies magnitude across each batched output, so the added computational complexity of computing cardinality augmented loss functions is minimal.

There are other rebalancing schemes and optimization methods which can be applied to account for class imbalance. SMOTE~\cite{SMOTE} is one such technique, taking linear subsamples to augment the majority class. Focal crossentropy~\cite{Focal} is another loss function based method, utilizing a transformation of the crossentropy loss curve to artificially suppress overrewarding easy classifications. Reweighting inputs based on class representation or explicitly undersampling the majority class are two additional options for addressing class imbalance, but both involve significant transformations to how the input data is handled and risk distorting the resulting neural network. Additionally, these methods could be applied in conjunction with cardinality augmented loss functions, as the application of our loss does not preclude other reweighting schemes where they are appropriate.

\subsection{Cardinality-like invariants}

We introduce some foundational definitions for the study of the invariants we make use of in this work. The two invariants we will examine are the magnitude, an isometric invariant of metric spaces introduced by Leinster~\cite{magnitude} and the spread, introduced by Willerton~\cite{spread}. The following definitions and observations are attributable to their work. For further information on magnitude, we direct the reader to Leinster's original work, as well as an appropriate survey paper~\cite{magnitudesurvey}.

\begin{definition}
    Let $(X,d)$ be a finite metric space. We refer to the matrix D such that \[ D_{ij}= d(x_i,x_j) \] as the \textbf{distance matrix} of $X$.
\end{definition}

\begin{definition}
    Let $(X,d)$ be a finite metric space, and $D$ the distance matrix of $X$. Then we refer to the matrix $\zeta_X$ such that \[ \zeta_{X_{ij}}= \mathrm{e}^{-d(x_i,x_j)} \] as the \textbf{similarity matrix} of $X$.
\end{definition}

The similarity matrix has entries which scale from 1 when two points are identical (that is, distance 0) and asymptotically approaches $0$ as $d(x_i,x_j)$ approaches infinity. In this sense, the similarity matrix can be considered an evaluation of how similar two points are.

\begin{definition}
    Let $(X,d)$ be a finite metric space and $\zeta_X$ its similarity matrix. If there exists a vector $\mathbf{w}$ such that $\zeta_X\mathbf{w} = \mathbf{1}$ then we say $\mathbf{w}$ is a \textbf{weighting} on $X$.
\end{definition}

\begin{definition}
    If $X$ admits at least one weighting, we say $X$ has \textbf{magnitude} and define \[|X| = \sum^{\#X}\mathbf{w}_i \] as the magnitude of $X$.
\end{definition}

\begin{example}
    Let $X$ be a space of two points distance $l$ apart. Then we can write the similarity matrix of $X$, \[\zeta_X=\begin{bmatrix}
				1& \mathrm{e}^{-l}\\
				\mathrm{e}^{-l}& 1
			\end{bmatrix}.\] Then, \[ \zeta_X^{-1}=\begin{bmatrix}
				\frac{1}{1-\mathrm{e}^{-2l}}& 	\frac{-\mathrm{e}^{-l}}{1-\mathrm{e}^{-2l}}\\
				\frac{-\mathrm{e}^{-l}}{1-\mathrm{e}^{-2l}}& \frac{1}{1-\mathrm{e}^{-2l}}
			\end{bmatrix}  \]
			So, \[ |X|=\sum_{x,y\in X}\zeta^{-1}_X(x,y)=\frac{2-2\mathrm{e}^{-l}}{1-\mathrm{e}^{-2l}}=\frac{2}{1+\mathrm{e}^{-l}}\]
\end{example}

\begin{figure}[t]
    \centering
    \includegraphics[width=1\linewidth]{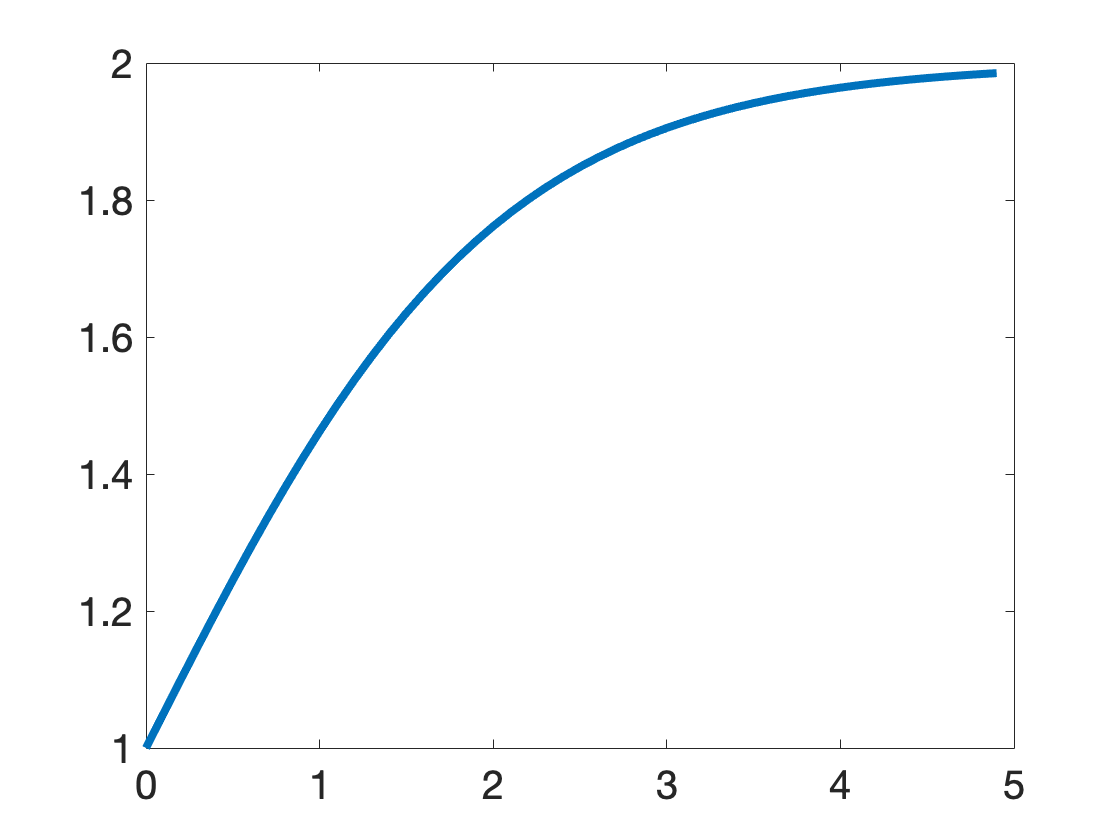}
    \caption{The magnitude function for the space of two points.}\label{fig:magnitude}
\end{figure}

We can similarly define the \textbf{spread} from this juncture, albeit without the need for a weighting.

\begin{definition}
    Let $(X,d)$ be a finite metric space and $\zeta_X$ its similarity matrix. We define \[ E_0(X) = \sum_i^{\#X}\frac{1}{\sum_j^{\#X}\zeta_{X_{i,j}}}  \] as the \textbf{spread} of $X$
\end{definition}

\begin{example}
    Let $X$ be a space of two points distance $l$ apart. Then we can write the similarity matrix of $X$, \[\zeta_X=\begin{bmatrix}
				1& \mathrm{e}^{-l}\\
				\mathrm{e}^{-l}& 1
			\end{bmatrix}.\] Then, \[ E_0(X)=\frac{2}{1+\mathrm{e}^{-l}}\]
\end{example}

Informally, the spread of $X$ may be defined as the sum of the inverse of the sum of the rows of the similarity matrix of $X$. Similar to magnitude, it scales between $1$ and $\#X$. Unlike magnitude, its computation is significantly cheaper, thus making the spread a more accessible computation if this is desired. 

To derive more information from the magnitude and the spread, we define the following functions.

\begin{definition}
    Let $(X,d)$ be a finite metric space. Then we denote by $tX=(X,d_{tX})$ the dilation of the metric space with pairwise distances $d_{tX}(x,y)=td(x,y)$ for $x,y\in X$.  We define the partially defined function 
    \begin{align*}
        (0,\infty)& \to\RR\\
        t&\mapsto |tX|
    \end{align*}
    to be the \textbf{magnitude function} of $X$.
\end{definition}

\begin{definition}
    Let $(X,d)$ be a finite metric space. Then we denote by $tX=(X,d_{tX})$ the dilation of the metric space with pairwise distances $d_{tX}(x,y)=td(x,y)$ for $x,y\in X$.  We define the partially defined function 
    \begin{align*}
        (0,\infty)&\to\RR\\
        t&\mapsto E_0(tX)
    \end{align*}
    to be the \textbf{spread function} of $X$.
\end{definition}

The following key properties of the magnitude function are results of Leinster~\cite{magnitude}.

\begin{proposition}[Leinster]
    Let $X$ be a finite metric space. Then \begin{enumerate}
        \item $|tX|$ is defined for all but finitely many $t>0$.
        \item The magnitude function $t\to |tX|$ is analytic wherever $\zeta_{tX}$ is invertible.
        \item There is some value $\hat{t}>0$ such that $\forall t>\hat{t}$, $|tX|$ is increasing and all weightings on $tX$ consist of exclusively positive terms.
    \end{enumerate}
\end{proposition}

We make special note of the analyticity property of magnitude, as this is desirable for loss functions in neural networks.

The following key properties of the spread are results of Willerton~\cite{spread}.

\begin{proposition}
    Let $X$ be a finite metric space. Then\begin{enumerate}
        \item $1\leq E_0(X)\leq \#X$
        \item $E_0(tX)$ is monotonically increasing w.r.t $t$.
        \item $E_0(tX)\to 1$ as $t\to 0$
        \item $E_0(tX)\to \#X$ as $t\to\infty$
        \item $E_0(X)\leq \mathrm{e}^{\text{diam}(X)}$
        \item $E_0(tX)$ is continuous and defined everywhere for $t\in[0,\infty)$
    \end{enumerate}
\end{proposition}

We note that the spread has a number of desirable properties for loss functions (monotonicity, continuity) which are immediately apparent. For magnitude, these are less obvious and require argument. The following statement is attributable to~\cite{magnitude},~\cite{extremalmagnitude}, and concurrently~\cite{mythesis}.

\begin{proposition}
    Let $X\subset\RR^n$ be a finite metric subspace of euclidean space, inheriting the euclidean metric. Then $|tX|$ is everywhere defined and continuous w.r.t $t$ for $t\in[0,\infty)$.
\end{proposition}

The following is conjectured, but holds in all known examples in practice.

\begin{conjecture}
    Let $X\subset\RR^n$ be a finite metric subspace of euclidean space, inheriting the euclidean metric. Then $|tx|$ increases monotonically w.r.t. $t$.
\end{conjecture}

\begin{definition}
    Let $y_{pred}$ and $y_{true}$ denote the output of a multiclass classifier neural network with batch size $b$ and number of classes $n$. Then we define the \textbf{magnitude loss} of the batched predictions as \[M(y_{pred}, y_{true}) = |\{y_{true}-y_{pred}\}\cup \mathbf{0}|-1, \] that is, the magnitude of the difference of the true and predicted classification as a one-hot vector with a $0$ vector appended to the end.
\end{definition}

\begin{definition}
    Let $y_{pred}$ and $y_{true}$ denote the output of a multiclass classifier neural network with batch size $b$ and number of classes $n$. Then we define the \textbf{spread loss} of the batched predictions as \[S(y_{pred}, y_{true}) = E_0(\{y_{true}-y_{pred}\}\cup \mathbf{0})-1, \] that is, the magnitude of the difference of the true and predicted classification as a one-hot vector with a $0$ vector appended to the end.
\end{definition}

The purpose of appending a zero vector to the magnitude computation is to ensure optimization converges around $0$, i.e, null loss. We subtract $1$ from the end since the minimal value attainable by the magnitude function is $1$ and it is desirable to have a loss function which attains $0$ at perfect classification.

We observe that the simplest way to compute magnitude is through the inversion of the similarity matrix, as this immediately provides an appropriate weighting, where possible. Certainly, one may note that matrix inversion is computationally expensive. However, as this complexity scales with the batch size, as long as batches are not exceptionally large, magnitude will not represent a noticable contribution to runtimes for training neural networks. Faster methods may involve solving the linear system for the weighting directly, or working through the Cholesky decomposition of $\zeta_X$.

We further note from our examples above that magnitude and spread loss are similar to Welsch-Leclerc loss~\cite{welsch}~\cite{leclerc} in the case of batch size 1. The WL loss can be defined as \[WL(y_{true},y_{pred}) = 1 - \mathrm{e}^{-\frac{1}{2} (y_{true}-y_{pred})^2}.\]   Both the Welsch-Leclerc loss and our loss functions rely on the negative exponential to scale loss observations, providing the additional benefit of suppressing outliers. We maintain this benefit, with the added advantage of normalizing for diverse loss observations across each batch. 

\section{Experiments}

We consider 3 scenarios for the implementation of magnitude in training neural networks. In the first two, we utilize the sklearn make classification function to produce an imbalanced artifical classification dataset with $10$ classes~\cite{scikit-learn}. This function follows work of Guyon~\cite{NIPS2003}. The classification creates 10 clusters around the vertices of a $15$ dimensional hypercube, with $15$ informative features and $5$ redundant features, which are linear combinations of the informative features. In the first dataset, we introduce a 50\% majority class with the remaining 50\% distributed among the other 9 classes, and in the second we introduce a 90\% majority class with the remaining 10\% distributed among the other classes. For training and evaluation, we apply a 7:3 train-test split.

\begin{table}[h]
    \centering
    \begin{tabular}{lcc}
    \hline
    \textbf{Feature} & \textbf{Dataset 1} & \textbf{Dataset 2} \\
    \hline
    \hline
    Samples & 10,000  & 10,000  \\
    Classes  & 10 & 10  \\
    Inf. feat. & 15 & 15 \\
    Red. feat. & 5 & 5 \\
    Maj. Class & 50\% & 90\% \\
    \hline
    \end{tabular}
    \caption{Composition of our two synthetic imbalanced datasets. The remaining non-majority class points are evenly distributed among the remaining minority classes.}
\end{table}

For our third scenario, we evaluate the performance improvement attainable through implementing cardinality augmented loss in a self-supervised learning environment for a two class classification task. The DeepGlassNet model~\cite{Glass} classifies glass composition types by their glass transition (GT) range, a critical indicator of material properties~\cite{glasstransition}. The DeepGlassNet authors employ a sanitized version of the v7.12 SciGlass dataset~\cite{sciglass}, an agglomerated dataset cataloguing various glass properties. The author's sanitization expunges samples with normalized mass summing to a value outside the range of $[.95,1.05]$, indicating data corruption or exceedingly imprecise measurement. We evaluate the dataset with a desired positive range of GT temperatures in the range $[500,600]$ with an 80/20 validation split, producing a training set with $8029$ positive samples and $20124$ negative samples, a roughly $72\%$ imbalanced negative majority class. The validation set contains $1942$ positive samples and $5081$ negative samples, for an imbalance proportional to the training set.

\begin{table}[h]
    \centering
    \begin{tabular}{lcc}
    \hline
    \textbf{Feature} & \textbf{Train} & \textbf{Valid} \\
    \hline
    \hline
    Pos. samples          &   8029           & 1942             \\
    Neg. samples          & 20124                 & 5081                 \\
    Maj. Class & 72\%     & 72\%               \\
    \hline
    \end{tabular}
    \caption{Composition of the DeepGlassNet dataset. The observations are segmented by membership in the positive or negative class, based on whether they melt within a given temperature range.}
\end{table}

\section{Methods}

For our synthetic datasets, we construct a simple neural network with a single fully connected hidden layer consisting of 32 units with ReLU activation. The output layer consists of a 10 unit FC layer with softmax activation. We train all NNs using SGD with a learning rate of .01, for 100 epochs and with a batch size of 32.

For the SciGlass dataset we mirror training procedure and network construction from DeepGlassNet. The DeepGlassNet model consists of a fully connected embedding layer, a graph convolution layer to characterize relationships between features in the input vector space, a self-attention layer, and a fully connected output layer. The graph convolution layer involves treating each feature of the input vector following embedding as a vertex on a graph and learning edge weights to determine pairwise relationships between features. The final output consists of a 1024 dimensional vector which is then compared through traditional shortest distance clustering with positive and negative anchor samples to determine class membership. For further information regarding the DeepGlassNet model we refer the reader to the DeepGlassNet paper~\cite{Glass}, as aside from the loss function our implementation is identical.

To modify the original DeepGlassNet network, we leave the structure of the original network unchanged and restrict our changes to the loss function applied during training. We introduce two new loss forumulations, which we term division spread and division magnitude. Let $S$ denote the full sample space, $P$ denote the set of positive observations, and $N$ denote the set of negative observations. The original network is trained over batches $B$ where 

\[ B = \{(s,p,n)\,|\, s\in S,\, p\in P,\, n\in N \} \]

so that each sample is paired with a positive and negative observation. Let $N$ denote the network and let $N(s)$ denote the output of the network for input $s$. Then the original loss function is defined as

\[ L_{orig}(N,B) = \frac{\sum_{(s,p,n)\in B}\log(\sigma(s,p,n,\tau))}{\#B}  \]

and 

\[ \sigma(s,p,n,\tau) = 1+\mathrm{e}^{-((N(s)\circ N(p))-(N(s)\circ N(n)))/\tau} \]

where $\circ$ denotes the Hadamard product and $\tau$ is a temperature hyperparameter. We provide two modifications to this loss function as follows:

\[ L_{dmag}(N,B) = \frac{L_{orig}(N,B)}{|\{s-n| (s,p,n)\in B  \}|} \]

and 

\[ L_{dspr}(N,B) = \frac{L_{orig}(N,B)}{E_0(\{s-n| (s,p,n)\in B  \})} \]

where $dmag$ and $dspr$ denote division magnitude and division spread, respectively. We report resulting metrics from all loss functions training the same network over 100 epochs, as well as progressions for these metrics.

All NNs are trained with the Adam optimizer, with a learning rate of $1e-6$ and an $L2$ penalty of $1e-7$. Networks are trained for $100$ epochs, with a batch size of $128$. Per the original work in DeepGlassNet, a small amount of noise is applied to each input vector to generate more varied inputs, demonstrating a scenario where cardinality augmented loss may be used cumulatively with other data augmentation methods.

\section{Results}

\subsection{Synthetic datasets}
For our synthetic datasets, we report precision-recall area-under-curve (PR-AUC), F1 score with micro and macro averaging, accuracy progression, and minimally observed loss values for CCE and MSE across other loss landscapes. For both imbalanced datasets, magnitude outperforms categorical crossentropy in both overall performance (accuracy, F1 micro) and performance on minority classes (F1 macro). Magnitude further confers no significant computational disadvantage, as the solving of such small linear systems is not a major contribution to runtimes.

Spread loss substantially underperforms compared to magnitude and categorical crossentropy on our synthetic datasets. This indicates that while spread shares many of the properties of magnitude, it is not suitable for use as a loss function in all situations, much like mean squared error (MSE). We further observe that MSE is in general unsuitable for imbalanced multiclass classification tasks, as the loss landscape trends far too dramatically towards the majority class.

We further report cross-performance across loss functions for CCE and MSE. We observe that optimization utilizing magnitude in our class imbalanced environment not only improves performance on the minority class, but further improves performance for other metrics and loss functions as well (CCE, MSE, see \ref{tab:imb50}). This implies that for this class imbalanced task, the loss landscape produced through magnitude loss is better suited to optimization for our classifier than the landscape generated through CCE. We thus conclude optimization through magnitude is preferable this scneario, and suggest it would be prudent to examine the performance advantage available through the application of magnitude as loss for other class imbalanced datasets. 

\subsection{DeepGlassNet}

\begin{figure*}[ht!]
        \centering
        
        \begin{minipage}{.45\linewidth}
            \includegraphics[width=\linewidth]{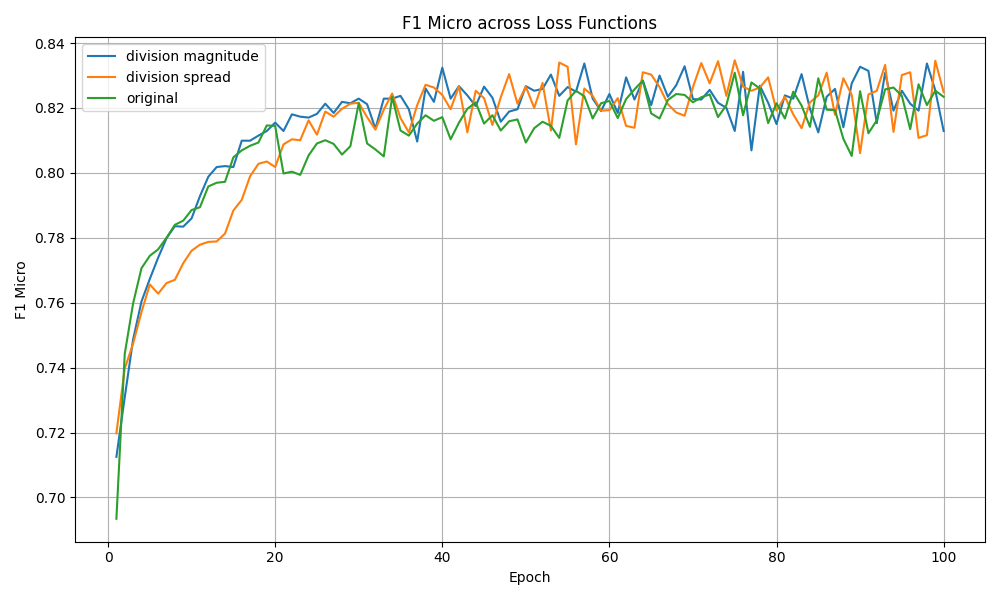}
        \end{minipage}
        \hfill
        \begin{minipage}{.45\linewidth}
            \includegraphics[width=\linewidth]{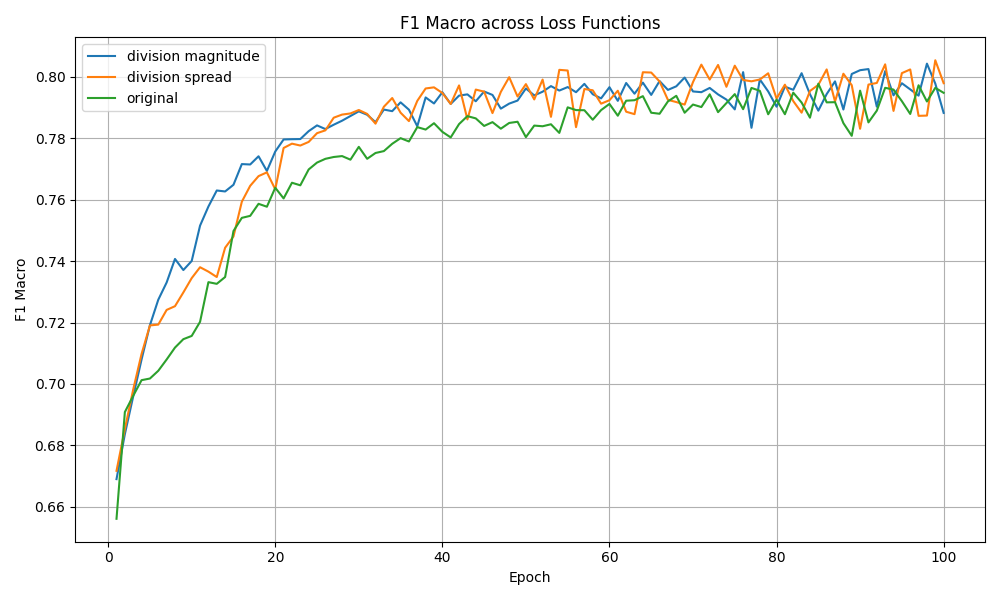}
        \end{minipage}
        
        \vspace{0.2cm}
        
        \begin{minipage}{.45\linewidth}
            \includegraphics[width=\linewidth]{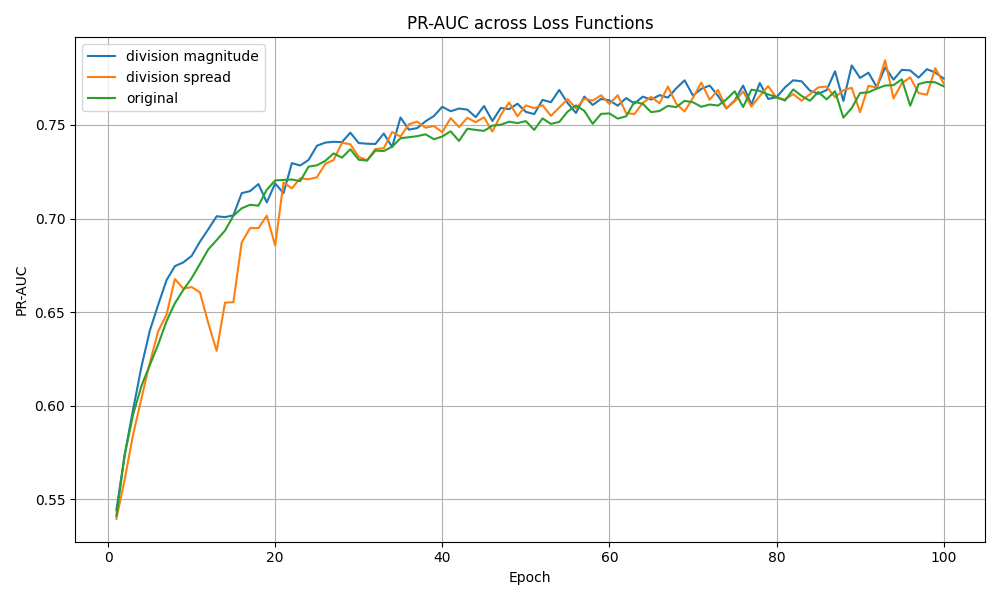}
        \end{minipage}
        \hfill
        \begin{minipage}{.45\linewidth}
            \includegraphics[width=\linewidth]{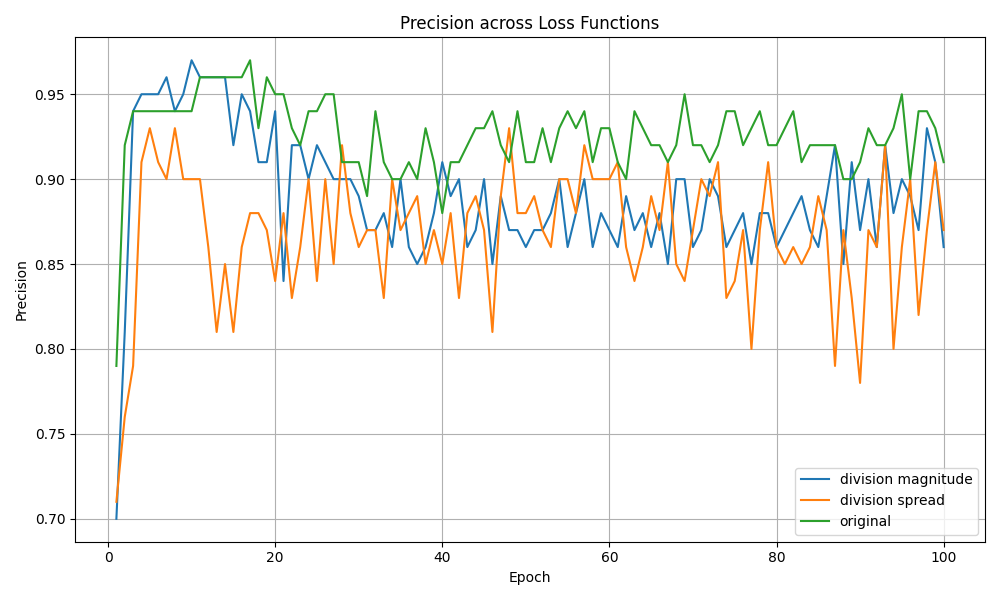}
        \end{minipage}
        
        \caption{Metric progression from the DeepGlassNet dataset.}\label{DeepGlassProg}
        
\end{figure*}

For the DeepGlassNet dataset, we report maximal F1 Macro, F1 Micro, precision, and PR-AUC scores~\ref{tab:DeepGlassMax} as well as progressions for these metrics~\ref{DeepGlassProg}. We observe that both magnitude and spread augmented loss confer improvement over the original loss function in all metrics aside from precision. For precision in particular, magnitude augmented loss preserves the precision score of the original loss function while conferring improvement in other metrics. Spread nominally produces the highest performance in all other metrics, but due to the loss in peak precision, we consider magnitude to be the preferable in this scenario.

\begin{table}[h]
    \centering
    \begin{tabular}{l|c|c|c|c}
    \hline
    \textbf{Met.} & \textbf{Mag.} & \textbf{Spr.} & \textbf{CCE} & \textbf{MSE} \\
    \hline
    \hline
    Acc.       & \textbf{.9090}                 & .8597             &.8857              & .5560   \\
    PR-AUC          & \textbf{.9860 }                & .9769             & .9839             & .8086    \\
    F1Macro        & \textbf{.8481 }              & .7646           & .8097             & .1530  \\
    Loss            & 1.3300                & .5584           & .3710             & .0538  \\
    CCE             & \textbf{.3338 }              & .4705           & .3710           & 1.3751  \\
    MSE             & \textbf{.0144  }             & .0202           & .0168             & .0538  \\
    \hline
    \end{tabular}
    \caption{Maximal performance for various loss functions on artificial classification task with 50\% majority class. Note Loss indicates the loss specified in the column title and is not comparable across results.}\label{tab:imb50}
\end{table}

\begin{table}[h]
    \centering
    \begin{tabular}{l|c|c|c|c}
    \hline
    \textbf{Met.} & \textbf{Mag.} & \textbf{Spr.} & \textbf{CCE} & \textbf{MSE} \\
    \hline
    \hline
    Acc.        & \textbf{.9557}                 & .9143             &.9440              & .8923   \\
    PR-AUC          & \textbf{.9539 }                & .9502             & .9517             & .8042    \\
    F1Macro        & \textbf{.6826 }              & .2605           & .6009             & .1114  \\
    Loss            & .7791                         & .2495           & .2260             & .0173  \\
    CCE             & \textbf{.2113}               & .3093           & .2260           & .4936  \\
    MSE             & \textbf{.0072}               & .0113           & .0091             & .0173  \\
    \hline
    \end{tabular}
    \caption{Maximal performance for various loss functions on artificial classification task with 90\% majority class. Note Loss indicates the loss specified in the column title and is not comparable across results.}\label{tab:imb90}
\end{table}

\begin{table}[h]
    \centering
    \begin{tabular}{l|c|c|c}
    \hline
    \textbf{Met.} & \textbf{Orig.} & \textbf{Spr.} & \textbf{Mag.} \\
    \hline
    \hline
    F1Micro.        & .8308             & \textbf{.8345}    &.8338             \\
    F1Macro        & .7978             & \textbf{.8051}             & .8044        \\
    PR-AUC        & .7743              & \textbf{.7847}           & .7818             \\
    Prec.            & .9700            & .9300           & .9700           \\
    \hline
    \end{tabular}
    \caption{Maximal performance for original, magnitude augmented, and spread augmented loss functions for the DeepGlassNet dataset. We observe that while spread nominally presents with the best result, the difference between spread and magnitude is not significant, while magnitude augmentation preserved the precision score of the original.}\label{tab:DeepGlassMax}
\end{table}

\begin{table}[h]
    \centering
    \begin{tabular}{lc}
        \hline
        \textbf{Loss} & \textbf{Avg s/epoch} \\
        \hline\hline
        Magnitude   & 0.4469 \\
        Spread  & 0.4049 \\
        CCE  & 0.4045 \\
        MSE   & 0.4311 \\
        \hline
    \end{tabular}
    \caption{Average time per epoch for various loss functions on artificial classification task with 50\% majority class. We note this computational time is taken from the tensorflow version, where optimization for our custom loss is better implemented.}
\end{table}

When training the DeepGlassNet network, we observe substantial improvement through the introduction of magnitude as a divisor to the existing metric. We suspect this improvement is attributable to gradient alignment for generating a more diverse classifier. In effect, the division of loss by the magnitude and spread causes loss values to decrease not only when the anchor and negative observation are distinct, but also when distinctions between anchors and negative values are distinct from each other. That is, the network is further rewarded for developing more diverse classifications instead of simply correct ones. We thus hypothesize the commensurate improvement in performance is attributable to the added robustness from developing a more diverse classifier.

\section{Discussion}

As might be expected from a method designed to address repetitious error attributable to class imbalance, the most substantial improvement conferred by cardinality augmented loss is observed among classifications of the minority. However, breaking from results observed in the work of Andreeva et. al.~\cite{A24}, we further observe improved performance of the network overall. In especially class imbalanced scenarios, magnitude further confers improved peak performance on CCE and MSE as metrics, even compared to optimization using those loss functions directly (See tables~\ref{tab:imb50},\ref{tab:imb90}).

\begin{figure*}[ht!]
    \centering
    
    \begin{minipage}{0.43\linewidth}
        \includegraphics[width=\linewidth]{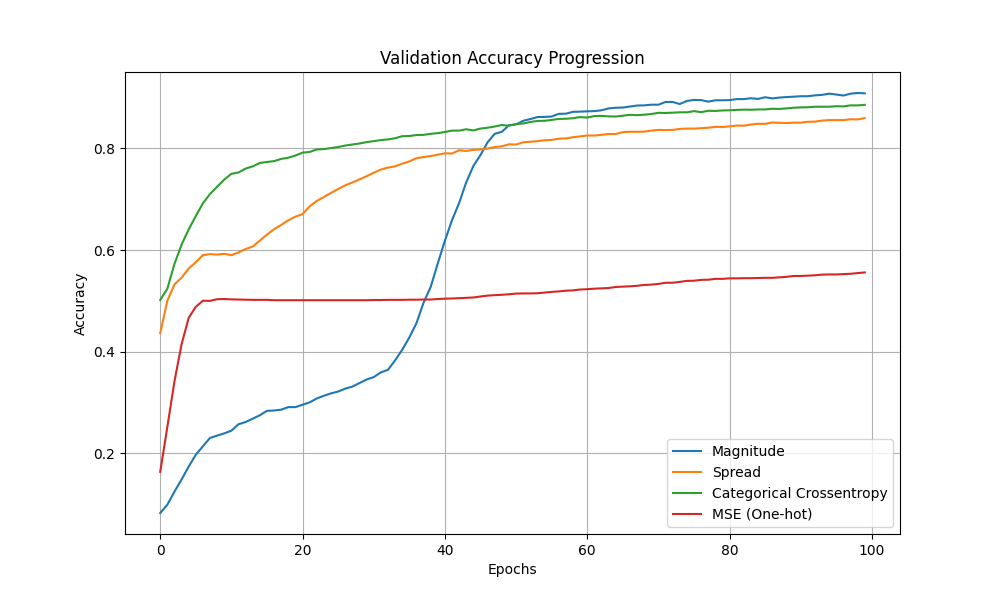}
    \end{minipage}
    \hfill
    \begin{minipage}{0.43\linewidth}
        \includegraphics[width=\linewidth]{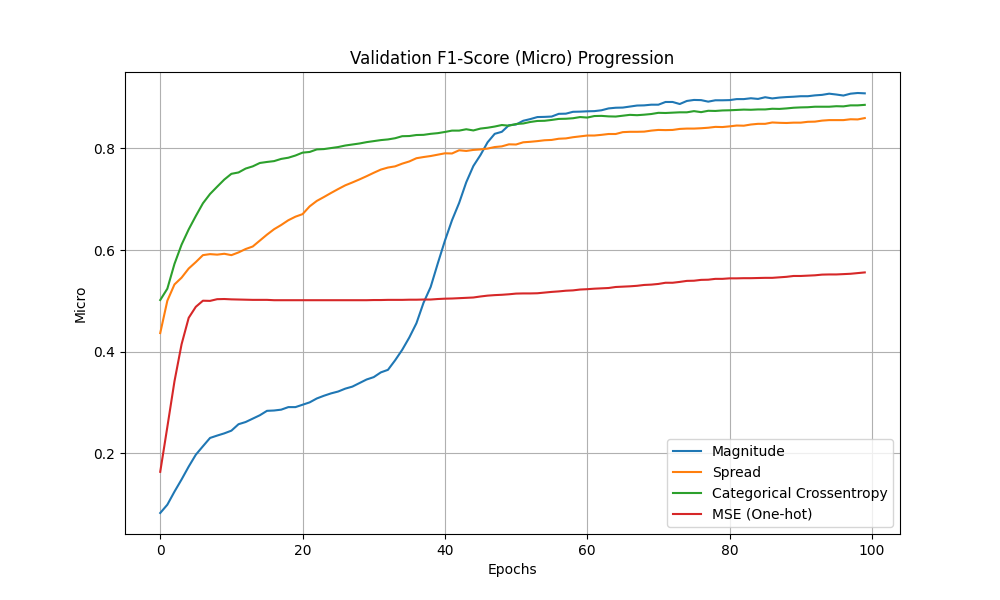}
    \end{minipage}
    
    \vspace{0.2cm}
    
    \begin{minipage}{0.43\linewidth}
        \includegraphics[width=\linewidth]{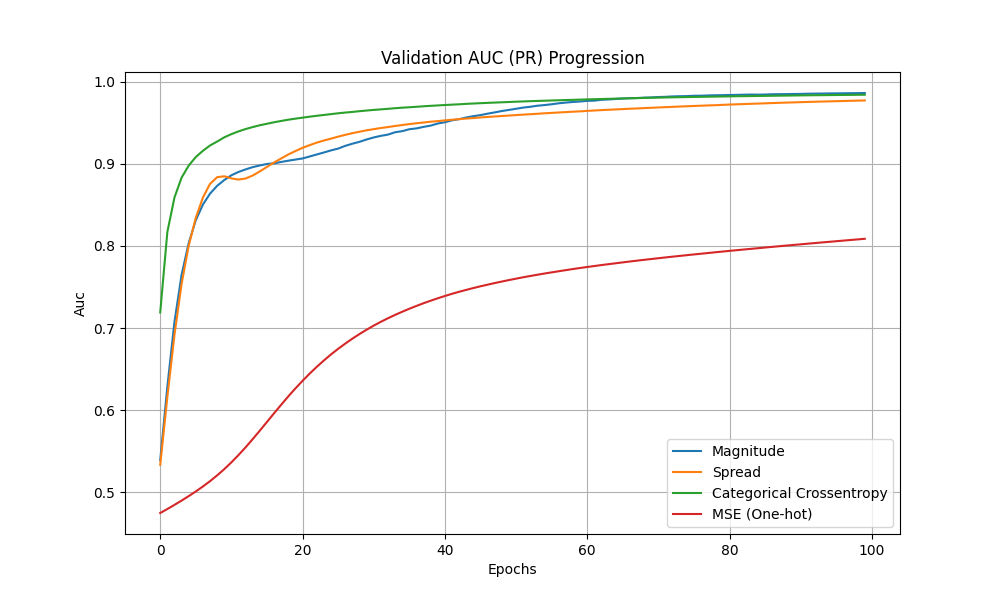}
    \end{minipage}
    \hfill
    \begin{minipage}{0.43\linewidth}
        \includegraphics[width=\linewidth]{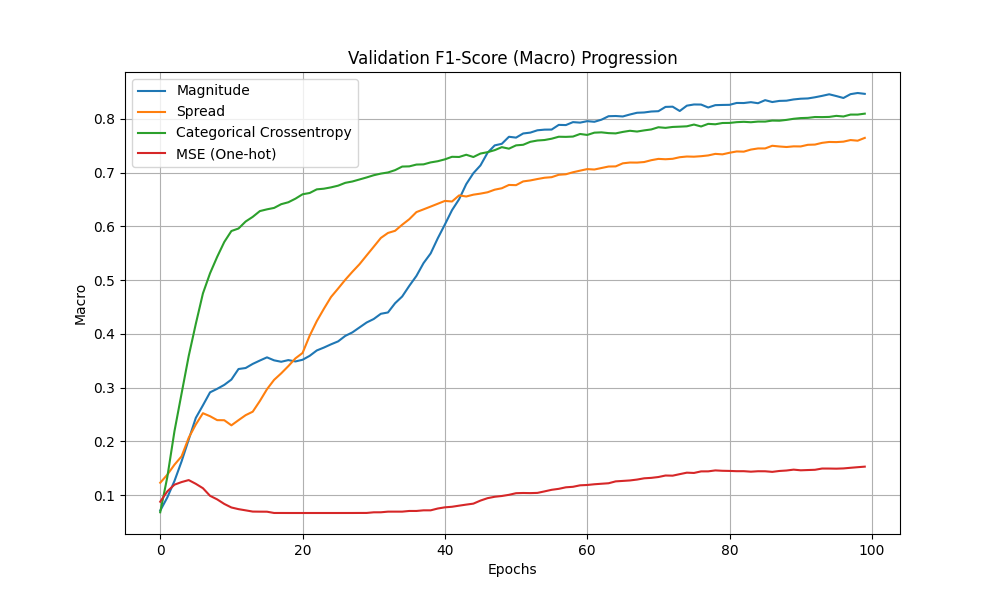}
    \end{minipage}
    
    \caption*{Results from the 50\% majority class synthetic dataset.}
    
    \vspace{0.5cm}
    
    \begin{minipage}{0.43\linewidth}
        \includegraphics[width=\linewidth]{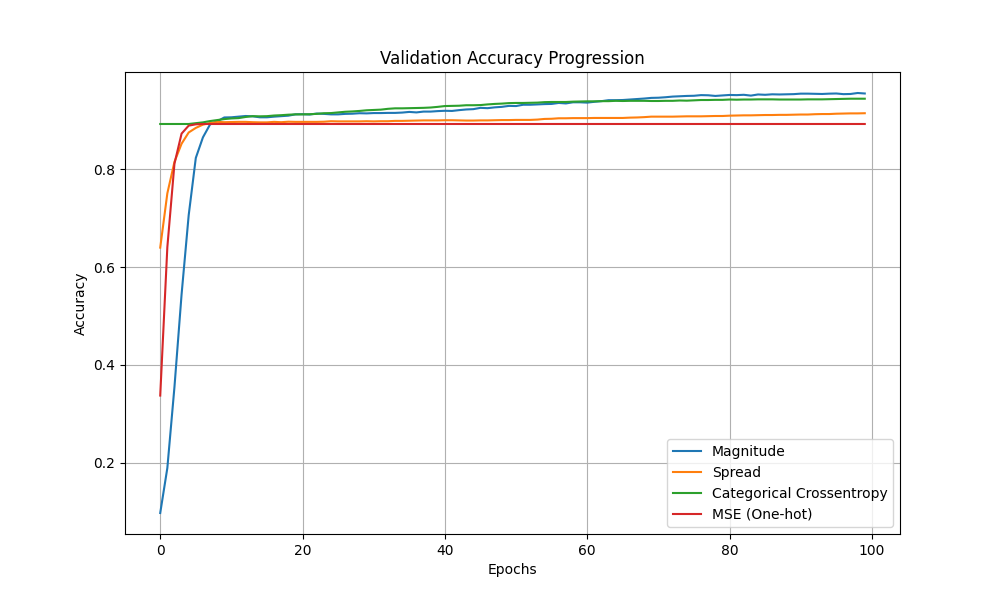}
    \end{minipage}
    \hfill
    \begin{minipage}{0.43\linewidth}
        \includegraphics[width=\linewidth]{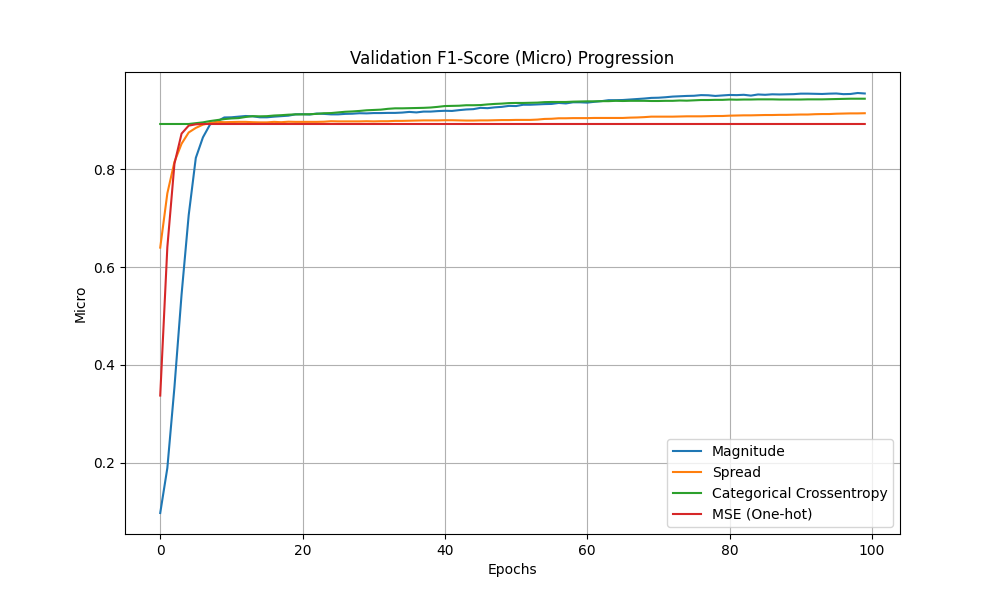}
    \end{minipage}
    
    \vspace{0.2cm}
    
    \begin{minipage}{0.43\linewidth}
        \includegraphics[width=\linewidth]{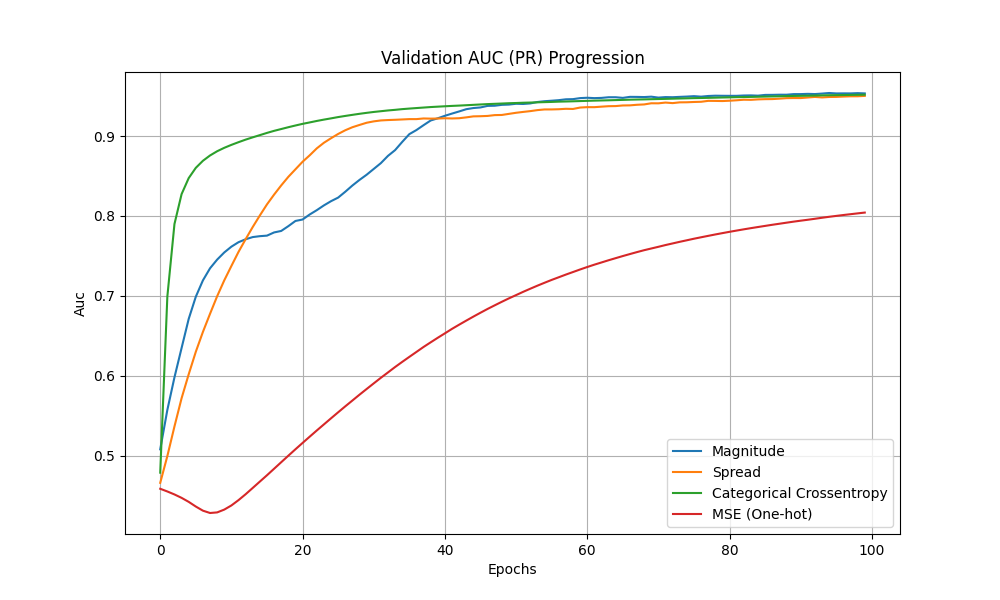}
    \end{minipage}
    \hfill
    \begin{minipage}{0.43\linewidth}
        \includegraphics[width=\linewidth]{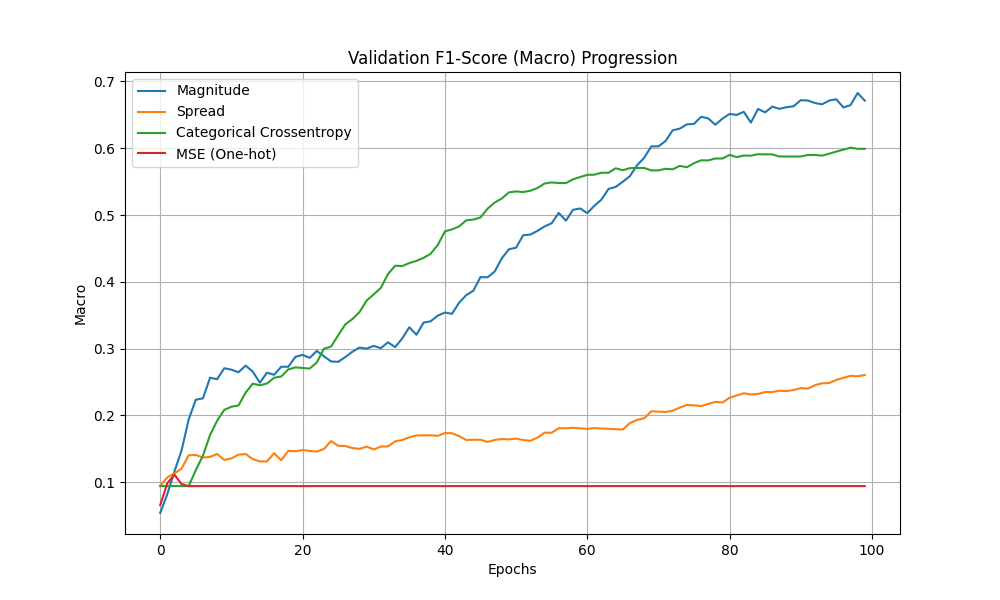}
    \end{minipage}
    
    \caption{Metric progression from the 50\% and 90\% majority class synthetic datasets.}\label{SynProg}
    \end{figure*}

Worth investigation is the significant startup period magnitude requires before surpassing the performance of other loss functions. As we can observe from Figure~\ref{SynProg}, magnitude has a significant warmup phase during which it underperforms even MSE before quickly ascending to surpass other loss functions. We suspect this behavior is largely attributable to the inherent bias magnitude loss has towards unique loss observations, a hypothesis further supported by a similar but significantly smaller warmup period observed for the spread. A further optimization for magnitude loss may be a weighted combination of magnitude and other loss functions along with a scheduling regime, which we leave for future work.

Of some note is the performance of division spread, in contrast to spread as a loss function. We observe significantly improved performance of the DeepGlassNet network through the application of both invariants, but the application of spread itself as a loss functions appears to be wholly unsuitable in the case of our synthetic datasets. Aside from higher peak precision for division magnitude, we observe no meaningful increase in performance through the application of magnitude over spread in this case. This may indicate that in this particular task spread is uniquely suitable, or it may indicate that for other imbalanced regimes where a cardinality-like invariant is desired, spread may be applicable as a computationally cheaper alternative to magnitude. We leave further characterization of this phemonomenon to future work.

We make note that computational time for magnitude augmented loss will increase substantially with larger batches. For especially large batches, the spread may be more suitable, depending on the task. However, so long as batches are not exceptionally large (2000 elements or more), magnitude does not appear to be a significant computational cost in the training loop. As spread only requires operations of low complexity, in no circumstances will it represent a significant computational commitment over other loss functions.

Magnitude is known to be stable (indeed, analytic~\cite{magnitude}) in subsets of euclidean space, wherein loss is computed under our construction. However, the convexity of magnitude is unknown. The monotonicity of magnitude relative to the scale factor for the underlying metric space is only conjectured at the time of writing. Proof for this property and for magnitude's convexity as a loss function would be desirable for this method's application. As it stands, however, no known counterexamples to these properties exist, and so in practice we consider our application to be relatively safe.

\section{Conclusions}

This work provides a methodology for the application of cardinality-like invariants as loss functions in neural network training. Through evaluation on both synthetic and a real-world class imbalanced dataset, we observe significant performance improvement through the application of cardinality augmented loss functions. Critically, our modifications to obtain this improvement were minimal, requiring only the substitution of the loss function in existing training loops. Further, our modifications are exclusively at the level of the loss function and scale with batch size, making even computationally expensive methods such as magnitude viable for production models. We provide peak performance metrics for all examined networks, as well as training trajectories, providing some insight into the performance characteristics of cardinality augmented loss functions. We additionally demonstrate the cumulative benefit of cardinality augmented loss applied in conjunction with other regularization methods through the improvement in performance observed compared to the original DeepGlassNet training methodology. In future work, we will seek to establish further theoretical foundations for the application of cardinality like invariants as loss functions. We will further seek other instances where the application of magnitude can improve neural network performance, in order to establish clear guidance for when such methods are appropriate. In particular, we will examine other instances where cardinaility like invariants can be applied without significantly impacting the computational complexity of training, such as vector quantization. 

\section{Code Availability}

Code for the reproducibility of these results can be found at this project's \href{https://github.com/miguelomalley/CardLoss}{github}.

\printbibliography

\end{document}